\documentclass{article}

\usepackage{microtype}
\usepackage{graphicx}
\usepackage{booktabs} 

\usepackage{hyperref}

\usepackage[preprint]{icml2024}

\usepackage{amsmath}
\usepackage{amssymb}
\usepackage{mathtools}
\usepackage{amsthm}
\usepackage{xcolor}
\usepackage{colortbl}

\usepackage[capitalize,noabbrev]{cleveref}

\theoremstyle{plain}

\theoremstyle{definition}

\theoremstyle{remark}

\usepackage[textsize=tiny]{todonotes}

\icmltitlerunning{DualFocus: Integrating Macro and Micro Perspectives in Multi-modal Large Language Models}

\usepackage{mmstyle}
\usepackage{float}
\usepackage{multirow}
\usepackage{subfig}

\newcommand{\bd}[1]{\textbf{#1}}
\newlength\savewidth
\newcommand{\tablestyle}[2]{\setlength{\tabcolsep}{#1}\renewcommand{\arraystretch}{#2}\centering\footnotesize}

\definecolor{Graylight}{gray}{0.9}

\begin{document}

\twocolumn[
\icmltitle{DualFocus: Integrating Macro and Micro Perspectives \\in Multi-modal Large Language Models}



\icmlsetsymbol{equal}{*}

\begin{icmlauthorlist}
\icmlauthor{Yuhang Cao}{cuhk,ailab}
\icmlauthor{Pan Zhang}{ailab}
\icmlauthor{Xiaoyi Dong}{ailab}
\icmlauthor{Dahua Lin}{ailab}
\icmlauthor{Jiaqi Wang}{ailab}
\end{icmlauthorlist}

\icmlaffiliation{cuhk}{The Chinese University of Hong Kong}
\icmlaffiliation{ailab}{Shanghai AI Laboratory}

\icmlcorrespondingauthor{Jiaqi Wang}{wangjiaqi@pjlab.org.cn}

\icmlkeywords{Machine Learning, ICML}

\vskip 0.3in

{
\begin{center}
    \centering
    \captionsetup{type=figure}
    \includegraphics[width=\textwidth]{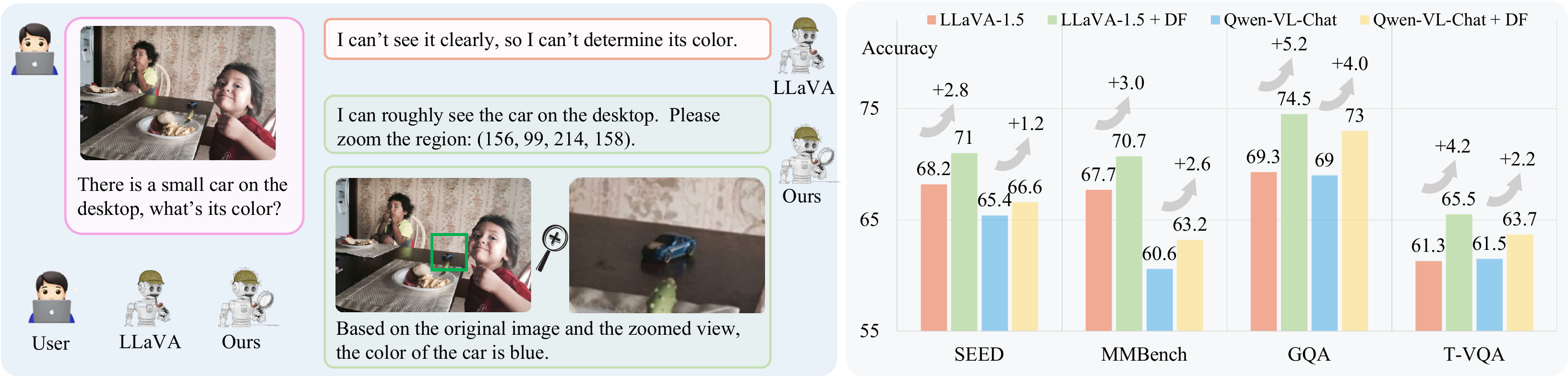}
    \vspace{-2mm}
    \captionof{figure}{
    Demonstrating the efficacy of DualFocus (DF) in enhancing multi-modal large language model (MLLM) performance. The left panel illustrates the scenario where a user asks an MLLM to identify the color of a small car in an image. Unlike the baseline MLLM (LLaVA), which struggles with detail, the DualFocus approach integrates an auto-zoom operation that precisely localizes and enlarges the area of interest. Consequently, DualFocus can accurately discern and report the car's color. The right panel corroborates DualFocus's superior performance, presenting a clear advantage in accuracy across multiple datasets (SEED, MMBench, GQA, T-VQA) compared to the baseline models (LLaVA, Qwen-VL). 
    }
    \label{fig:teaser}
    \vspace{3mm}
\end{center}
}

]



\printAffiliationsAndNotice{} 

\begin{abstract}
We present DualFocus, a novel framework for integrating macro and micro perspectives within multi-modal large language models (MLLMs) to enhance vision-language task performance. 
Current MLLMs typically singularly focus on inputs at a predefined resolution, resulting in deficiencies in detailed questions involving local regions.
We introduced a DualFocus mechanism where the model concentrates on the image from a macro perspective, responds to the question, and identifies suitable sub-regions to zoom in for subsequent micro perspective analysis. Via the integration of answers from both macro and micro perspectives, the model is adept at addressing tasks that encompass global, detailed, and combined considerations.
To endow the DualFocus mechanism in MLLMs, we curated a tailored dataset derived from the Visual Genome (VG)  and adapted it to align with the training regimen of DualFocus. 
Through comparative studies across different model sizes and benchmarks, we demonstrate DualFocus's superiority in balancing detailed examination with holistic insight, significantly reducing hallucination instances in MLLMs and improving their performance in various vision-language tasks.
Code is available at: \url{https://github.com/InternLM/InternLM-XComposer/blob/main/projects/DualFocus}
\end{abstract}

\section{Introduction}
\label{sec:introduction}

Large Language Models (LLMs) like ChatGPT \cite{openai2020chatgpt}, GPT-4 \cite{openai2023gpt4}, and PaLM \cite{chowdhery2022palm} have revolutionized the field of natural language processing with their astounding ability to follow human instructions and tackle open-ended tasks. These models demonstrate an exceptional understanding of language and can generate text that is often indistinguishable from that produced by humans.
Building upon this foundation, Multi-modal Large Language Models (MLLMs) such as MiniGPT-4 \cite{zhu2023minigpt}, LLaVA \cite{liu2023visual}, and InstructBLIP \cite{dai2023instructblip} have emerged, integrating the linguistic prowess of LLMs with visual understanding capabilities. Drawing on open-source LLMs like LLaMA \cite{touvron2023llama}, Qwen \cite{qwen7b} and InternLM \cite{2023internlm}, these MLLMs extend their insight to the visual domain, allowing for a more comprehensive understanding of questions that necessitate both visual and textual processing.

One of the primary challenges in advancing MLLMs resides in effectively incorporating visual information.
Current models such as MiniGPT-4 \cite{zhu2023minigpt} and LLaVA \cite{liu2023visual} often rely on imagery of a fixed, small resolution. This approach simplifies processing but limits the model's ability to discern micro details crucial for answering specific questions.
%
Conversely, models like Monkey \cite{li2023monkey} with 896$\times$896 pixel inputs and OtterHD \cite{li2023otterhd} with 1024$\times$1024 pixel inputs leverage high-resolution images to tackle fine detailed visual analysis. 
However, dramatically increased image resolution introduces overwhelming details, where most of them are irrelevant to a specific question. As a result, it is more challenging to attain a balance between global context and local information, which is crucial to benchmarks that demand a more holistic understanding like MMBench \cite{MMBench} and SEED~\cite{seed_2023}.

Drawing inspiration from human cognitive processes where an individual typically scans an image globally before focusing on relevant details to answer a question, we argue for the integration of similar strategies in MLLMs. 
Unlike existing approaches that follow a SingleFocus strategy, which looks the image once and answer the question directly.
Our proposed DualFocus strategy imitates human cognition by first analyzing the entire image to grasp the macro context, thereby deducing the important areas. Subsequently, the model zooms into these identified subareas for a detailed examination to accurately answer the provided question.
This method, inspired by the Chain of Thought (CoT) \cite{wei2022chain} in NLP, is extended by intertwining visual cues into the cogitative sequence through an auto-zoom mechanism.

To endow MLLMs with the DualFocus capacity to tackle both the micro and macro perspectives within an image, 
we curate a new dataset that derived from the Visual Genome (VG) \cite{krishna2017visual}, where we meticulously select images and annotations and explicitly align the format with our dual focus protocol.  
Throughout the model training phase, the MLLM learns to discern the relevant coordinates defining the important subregion for any given query. The model then leverages a magnified view of this localized area to ascertain the correct answer. 
During the inference stage, the model employs macro and micro answer pathways, respectively, thereby yielding two potential answers. To select the optimal response, we employ Perplexity (PPL) \cite{jelinek1998statistical} as a decision metric, comparing the computed losses from the two answers and opting for the one with the lower loss as the final prediction.

In our experimental evaluation, we utilize LLaVA 1.5 \cite{llava1_5} and Qwen-VL-Chat \cite{bai2023qwen} as baseline models, selected for their robust performance in multi-modal tasks.
Comparative experiments were conducted across model sizes of 7B and 13B parameters, and a diverse set of benchmarks that ranged from multi-modal and traditional VQA benchmarks.
Specifically, DualFocus improves LLaVA 1.5 by 2.8, 3.0, 4.2, 4.2 and Qwen-VL-Chat by 1.2, 2.6, 4.0, 2.2, on SEED \cite{seed_2023}, MM-Benchmark \cite{MMBench}, TextVQA \cite{singh2019towards}, and GQA \cite{hudson2018gqa}, respectively.
Additionally, we observed a notable reduction in hallucinatory responses in MLLMs when tested on the POPE benchmark \cite{li2023evaluating}, highlighting the framework's potential to curb the generation of spurious detail by maintaining a balanced perspective. 
The comparative studies reinforce the versatility of DualFocus across a spectrum of datasets, affirming the effectiveness of the DualFocus mechanism. 
\section{Related Work}
\subsection{Large Language Model (LLM)}
The evolution of LLMs has significantly shaped the natural language processing (NLP) landscape, showcasing the extraordinary capabilities of the Transformer architecture. Initiated by encoder-decoder models such as BERT \cite{devlin2018bert}, T5 \cite{raffel2020exploring} and decoder-centric architectures like GPT \cite{openai2020chatgpt}, these models have excelled across various NLP tasks. With GPT3 \cite{brown2020language}, decoder-only models have become increasingly prevalent due to their effectiveness in few-shot and zero-shot scenarios. Enhancements in model parameterization and dataset breadth are epitomized by Google’s PaLM \cite{chowdhery2022palm}, which pushed the performance boundaries of LLMs even further. To tailor models for natural conversational responses, strategies such as fine-tuning and reinforcement learning derived from human feedback have been adopted in InstructGPT \cite{ouyang2022training} and ChatGPT \cite{openai2020chatgpt}. The contribution of the open-source community, exemplified by the release of models like LLaMA \cite{touvron2023llama}, Vicuna \cite{vicuna2023}, Qwen \cite{qwen7b}, LLaMA2 \cite{touvron2023llama2}, Baichuan2 \cite{baichuan2023baichuan2}, and InternLM \cite{2023internlm}, has spurred a continuous stream of innovation, setting new benchmarks for NLP research.

\subsection{Multi-Model Large Language Model (MLLM)}
Recent research in MLLM has made significant advances in the visual language learning domain by exploring the integration of visual knowledge into LLMs. 
Models such as CLIP \cite{radford2021learning,sun2023alpha} and BLIP \cite{li2022blip} have demonstrated the effectiveness of contrastive learning to synchronize image and text modalities, remarkably improving zero-shot learning in tasks like Image Captioning and Image-Text Retrieval.
Models such as MiniGPT-4 \cite{zhu2023minigpt}, LLaVA \cite{liu2023visual}, InstructBLIP \cite{dai2023instructblip}, and Otter \cite{li2023otter} have pushed further, enhancing dialogic interactions and contextual understanding in image-text scenarios by focusing on precise pre-training alignments and fine-tuning processes. 
Notably, advanced techniques employing grounding data have been developed to anchor the models' perceptions more firmly in reality, as demonstrated by mPLUG-Owl \cite{ye2023mplug}, Shikra \cite{chen2023shikra}, Opera \cite{huang2023opera}, VIGC \cite{huang2023opera} and KOSMOS-2 \cite{peng2023kosmos}. Such initiatives mitigates the issue of hallucinations and leads to more reliable performances across visually grounded tasks, together with more rich multi-modality datasets \cite{zhao2023mllm,He2023WanJuanAC,wang2023v3det} resulting in the development of more advanced MLLMs \cite{internlmxcomposer,internlmxcomposer2,hong2023cogagent,qi2023gpt4point,qi2023gemini}.

\begin{figure*}[ht]
    \centering
    \includegraphics[width=\linewidth]{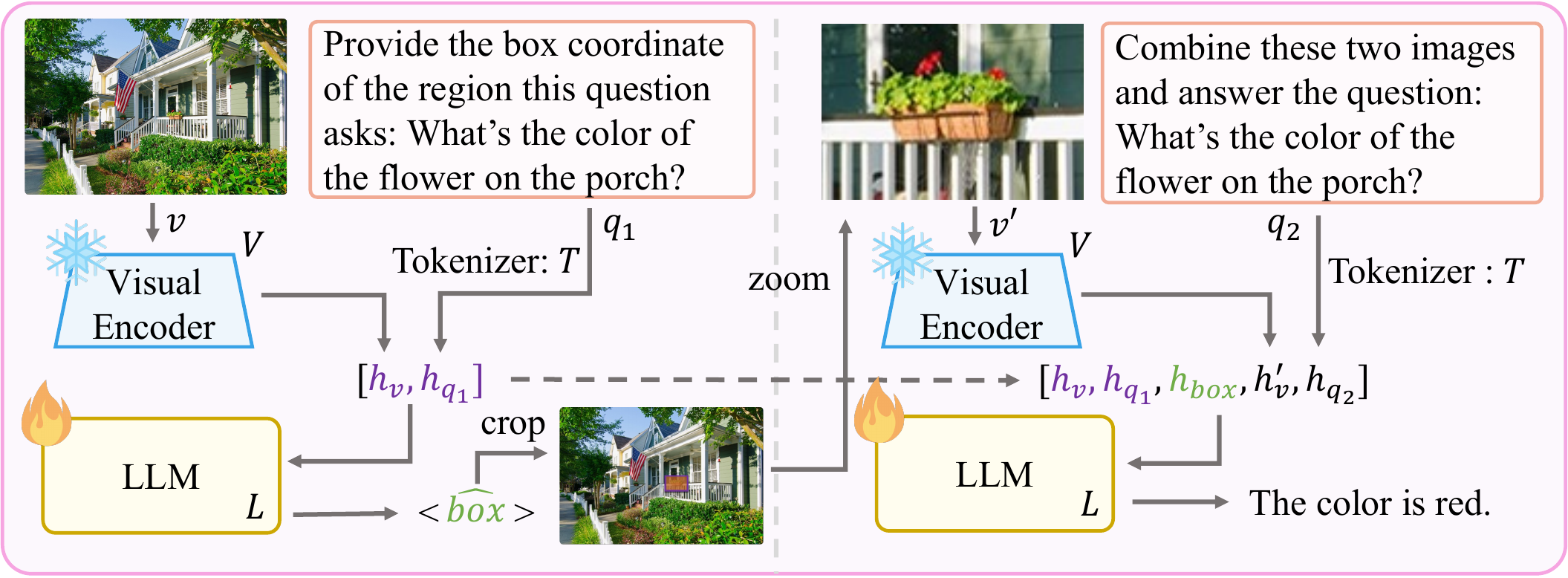}
    \caption{\bd{Framework.} Initial processing involves feature extraction from a visual input ($v$) via Visual Encoder ($V$) and question tokenization ($q_1$) via tokenizer $T$ to determine the subregion $\hat{box}$ (for brevity we omit connection layer $W$ here), followed by the generation of a localized, zoomed view ($v'$). The MLLM then fuses the macro-contextual representation ($h_v, h_{q_1}, h_{box}$) with the micro-focused information ($h_{v'}, h_{q_2}$) to accurately respond to the query ($\hat{ans}$). 
    }
    \label{fig:framework}
\end{figure*}

\subsection{High Resolution MLLMs}
Recently, MLLMs have primarily utilized fixed, lower-resolution inputs, typically 224 pixels \cite{liu2023visual,chen2023shikra,zhu2023minigpt}. Innovations such as LLaVA-1.5 \cite{llava1_5} and BLiVA \cite{Hu2023BLIVAAS} have sought to enhance performance by expanding input resolution to 336 pixels and integrating task-specific with global features, respectively. Moreover, advancements like Qwen-VL \cite{bai2023qwen} and OtterHD \cite{li2023otterhd} have pushed resolution boundaries to 448 pixels and preserved original image sizes during inference, leading to more refined detail discernment. Notably, Monkey \cite{li2023monkey} has significantly increased resolution to 896 pixels. Nonetheless, an inherent tension persists within MLLM architectures: models employing lower resolution inputs struggle with detecting finer details, whereas those with higher resolutions often underperform on tasks emphasizing global comprehension.
This paper introduces the DualFocus mechanism, reconciling the conflicting demands of micro-detail reflectivity and macro-contextual understanding, thereby offering a nuanced equilibrium for MLLM designs.

\section{Our Approach}
\label{sec:method}

In this section, we provide an initial overview of the Multi-modal Large Language Model (MLLM) (Sec. \ref{sec:recap_llava}). Following that, we elucidate the methodology, covering aspects such as dataset construction (Sec. \ref{sec:data_construction}), the training phase (Sec. \ref{sec:training}), and the inference process (Sec. \ref{sec:inference}).

\subsection{Preliminaries}
\label{sec:recap_llava}

The contemporary MLLMs usually adopt a modular architecture, comprising a visual encoder \(V\), a series of connection layers \(W\), and a large language model \(L\). Given an input image \(v\) and its corresponding question \(q\), the visual encoder \(V\) initially processes the image and encodes it into a set of visual tokens \(z_v = V(v)\). These visual tokens are then transformed to align with the embedding space of the language model through the connection layers, such that \(h_v = W(z_v)\). Concurrently, the text query \(q\) is tokenized into linguistic tokens \(h_q\) by the tokenizer \(T\), becoming \(h_q = T(x_q)\). These visual and text tokens are concatenated into a unified sequence \([h_v, h_q]\), which serves as the input to the decoder component of the large language model \(L\). The model then utilizes this combined representation to infer the appropriate answer \(ans = L([h_v, h_q])\), demonstrating the capability of these models to perform cross-modal reasoning and answer multimodal queries.

\subsection{Data Construction}
\label{sec:data_construction}
To enhance the MLLM with the DualFocus mechanism, we curated a dataset derived from the extensive Visual Genome (VG) dataset \cite{krishna2017visual}, which provides a diverse array of images coupled with corresponding questions, answers, and annotated bounding boxes. These bounding boxes explicitly demarcate the regions of interest within the image pertinent to the question posed.

\bd{Ambiguity Filtration}.
Initially, we scrutinize each data entry from VG to ensure its precision and clarity. During this process, we encountered instances where a question such as ``What is the color of the person's shirt?" might correspond to a scene depicting multiple individuals, leading to ambiguity in the dataset. To establish a one-to-one mapping between visual cues and textual queries, we employed a strict filtering criterion to exclude such ambiguous samples. Through this rigorous refinement, we distilled our dataset to 143k unequivocal image-question pairs.

\bd{Reformatting}.
For enhanced interaction with our MLLM's training regime, we transmuted the dataset samples into a conversational format that encapsulates both the query and spatial awareness components. The schema of a data sample is as follows,
\vspace{-2mm}
\begin{align*}
& \texttt{$Q_1$}: \texttt{<img>} \textit{Provide the box coordinates of the region} \\
& \quad \quad \textit{this question is asking about:} \enspace \texttt{<question>} \\
& \texttt{$A_1$}: \texttt{<box>} \\
& \texttt{$Q_2$}: \texttt{<sub img>} \textit{Combine these two images and} \\ & \quad \quad \textit{answer the question:} \enspace \texttt{<question>} \\
& \texttt{$A_2$}: \texttt{<answer>} 
\vspace{-2mm}
\end{align*} 
In the first round of inquiry (\bd{$Q_1, A_1$}), we task the MLLM with deducing the important subregion \texttt{<box>} that pertinent to the question \texttt{<question>} in the image \texttt{<img>}, supplying it with micro-level details it needs to focus on.
The subsequent round (\bd{$Q_2, A_2$}) is constructed to aggregate the augmented view \texttt{<sub img>} of the identified sub-region and the original contextualized image \texttt{<img>} to infer the answer \texttt{<answer>}.

\subsection{Training}
\label{sec:training}
During training, we integrate our curated VG data with standard VQA datasets to enhance the model's capabilities on both micro and macro levels.
We adhere conventional MLLM training procedures using standard VQA datasets to equip the model with macro capabilities.
Subsequent sections primarily focus on elaborating how we augment the model's proficiency in identifying micro details through our transformed VG data.
This enhancement is achieved by dividing the training process into two distinctive yet interconnected tasks.

\bd{Task \uppercase\expandafter{\romannumeral1}: Identification of the Pertinent Subregion}.
Given an image $v$ and the query $q$, we prompt the model with instruction $q_1$ to identify the region corresponding to the query $q$. To model this, we tokenize $q_1$ into tokens $h_{q_1}$ using the tokenizer $T(.)$, and the visual embedding $h_v$ is obtained from the input image $v$. The model prediction $\hat{box}$, representing the bounding box coordinates, is then inferred through the language model: 
\begin{equation}
   \hat{box} = L([h_v, h_{q_1}]),
\end{equation}
where $\hat{box} = (\hat{x_1}, \hat{y_1}, \hat{x_2}, \hat{y_2})$, representing the coordinates of the two corners of the bounding box within the image. The coordinates are expressed as numeric values embedded in natural language, with no additional formatting or special tokens, to maintain coherence with the LLM's language processing capabilities.

\bd{Task \uppercase\expandafter{\romannumeral2}: In-depth Examination and Answer Generation}.
Upon deducing the focused region, we extract and upscale the sub-image $v'$ using the corresponding bounding box coordinates to maintain the original resolution: 
$v' = zoom(crop(v, box))$. 
To ensure that the context of the entire image is not lost, both the original image $v$ and the processed sub-image $v'$ are encoded separately by the visual encoder, producing two sets of visual tokens $h_v$ and $h_v'$, respectively.
These visual tokens are concurrently concatenated with the text embedding generated from the original question and its corresponding answer from the first task, structured as $[h_v, h_{q_1}, h_{box}, h_v', h_{q_2}]$. The model then employs this concatenated information to produce the final answer,
\vspace{-2mm}
\begin{equation}
  \hat{ans} = L([h_v, h_{q_1}, h_{box}, h_v', h_{q_2}]).
  \label{equ:attend}
\end{equation}
\bd{Objective Function}.
The training loss is partitioned into two distinct segments corresponding to the tasks detailed above. Since both the bounding box and the final answer are enunciated in natural language, we employ a standard cross-entropy loss function $\mathcal{L}_{CE}$ for each task. Formally, the collective loss is the aggregation of these binary components:
\vspace{-1mm}
\begin{equation}
  \mathcal{L}_{total} = \mathcal{L}_{1}(\hat{box}, box) + \mathcal{L}_{2}(\hat{ans}, ans),
\end{equation}
where $\mathcal{L}_{1}$ computes the discrepancy between the actual ($box$) and predicted ($\hat{box}$) bounding boxes, and $\mathcal{L}_{2}$ quantifies the differential between the true final answer ($ans$) and the inferred one ($\hat{ans}$).

Through systematic training across these two focused tasks, the model gradually develops an adeptness in isolating and scrutinizing specific subregion within an image, thereby refining its capacity for fine-grained detail discernment.

\begin{figure}[t]
    \centering
    \includegraphics[width=\linewidth]{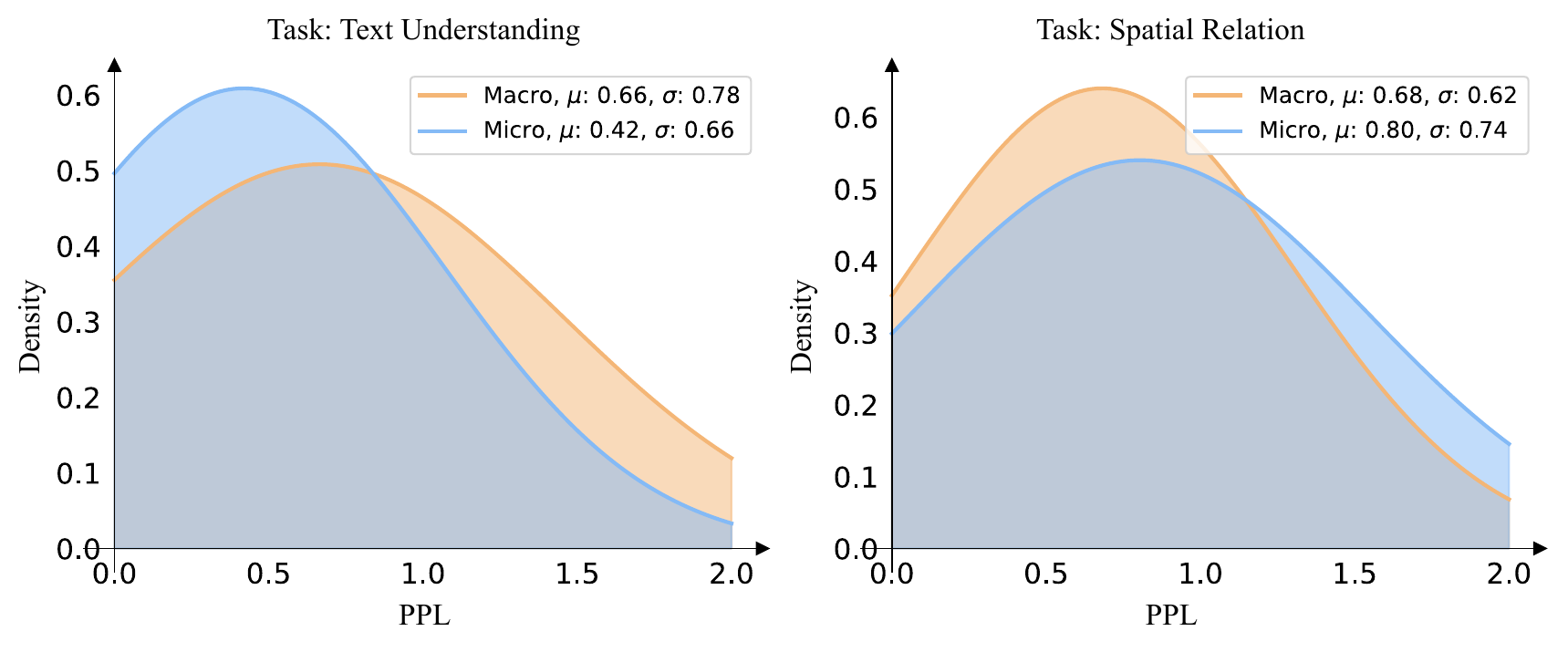}
    \caption{PPL distribution for MicroFocus compared to baseline on tasks emphasizing different cognitive demands.}
    \label{fig:ppl}
\vspace{-2mm}
\end{figure}

\subsection{Inference}
\label{sec:inference}

Upon training completion, our model acquires dual capabilities, namely the ability to generate macro-level answers (\(\hat{ans}_{\text{macro}}\)) directly from the holistic image and the capacity to produce micro-level answers (\(\hat{ans}_{\text{micro}}\)) using the fine-grained details from the predicted subregion. Thus, we adopt two distinct pathways for interpreting the given data.

\bd{Inference Pathways}.
Specifically, the macro answer pathway, akin to the traditional method, which maintains the conventional inference process, directly generating an answer, 
\begin{equation}
    \hat{ans}_{\text{macro}} = L([h_v, h_q]),
\end{equation}
without emphasizing localized regions.
Contrarily, the micro answer pathway mimics the training phase,
\begin{equation}
   \hat{ans}_{\text{micro}} = L([h_v, h_{q_1}, h_{\hat{box}}, h_v', h_{q_2}]), 
\end{equation}
except during inference we utilize the predicted bounding box $\hat{box}$ instead of the ground truth in Equ. \ref{equ:attend}. The micro pathway leverages the predicted bounding box $\hat{box}$ to focus on a specific sub-region.

\bd{Answer Selection Via Perplexity}.
To ascertain the most coherent response, we evaluate both $\hat{ans}_{\text{macro}}$ and $\hat{ans}_{\text{micro}}$ through their respective perplexity (PPL). The PPL serves as an estimate of the likelihood for a given sequence of tokens, with lower values indicating higher probability (better model confidence). This is given by:
\begin{equation}
    \text{PPL}(\hat{ans}) = \exp\left(-\frac{1}{N}\sum_{i=1}^N \log{p(w_i | w_{<i})}\right),
\end{equation}
where $N$ is the number of tokens in the answer, $p(w_i | w_{<i})$ represents the model's estimated probability for token $w_i$ given the preceding context. 
The answer affiliated with the lower PPL is deemed more likely correct and thus selected as the final answer:
\begin{equation}
  \hat{ans} = \begin{cases}
  \hat{ans}_{\text{macro}}, & \text{if PPL}(\hat{ans}_{\text{macro}}) < \text{PPL}(\hat{ans}_{\text{micro}}) \\
  \hat{ans}_{\text{micro}}, & \text{otherwise}
  \end{cases}
\end{equation}

\bd{The Motivation of Perplexity-guided Selection}.
As depicted in Fig. \ref{fig:ppl} (Left), the micro answer demonstrates superior confidence in scenarios requiring detailed discernment (\eg, text understanding). However, its performance degrades in tasks involving global comprehension (\eg, spatial relationships), as shown in Fig. \ref{fig:ppl} (Right), despite the micro answer being generated by concatenating the original image and sub-images. We conjecture that this degradation is attributable to the high dependence of the micro answer on the nearest image, i.e., sub-image, akin to the high dependency observed in closely located text tokens.
This suggests selecting the micro and macro answers via perplexity to integrate both perspectives during the inference of MLLM.
By using a perplexity-guided dual-path inference system, the MLLM dynamically switches between a global understanding and a focused comprehension dependent on the nature of the query, ultimately enhancing the efficacy of the model’s responses.

\begin{table*}[h]
    \centering
    \caption{\bd{Comparison with baseline methods on various benchmarks. }Our DualFocus consistently demonstrates improvements across various baselines and benchmarks.}
    \label{tab:main_base}
    \begin{tabular}{l|c|c|c|c|c|c}
    Method & Encoder-V & LLM & SEED$^{IMG}$ & MMB & GQA$^{*}$ & VQA$^{T}$ \\
    \midrule
    LLaVA-1.5    & \multirow{2}{*}{ViT-L} & \multirow{2}{*}{Vicuna-7B}  & 66.2 & 64.3 & 67.2 & 58.2 \\
    \bd{LLaVA-1.5 + DF} &         &                    & \bd{68.9 (+2.7)} & \bd{66.6 (+2.3)} & \bd{69.3 (+2.1)} & \bd{62.0 (+3.8)} \\
    \midrule
    LLaVA-1.5    & \multirow{2}{*}{ViT-L}  & \multirow{2}{*}{Vicuna-13B} & 68.2 & 67.7 & 69.3 & 61.3 \\
    \bd{LLaVA-1.5 + DF} &              &               & \bd{71.0 (+2.8)} & \bd{70.7 (+3.0)} & \bd{74.5 (+4.2)} & \bd{65.5 (+4.2)} \\
    \midrule
    Qwen-VL-Chat   & \multirow{2}{*}{ViT-G}   & \multirow{2}{*}{Qwen-7B} & 65.4 & 60.6 & 69.0 & 61.5 \\
    \bd{Qwen-VL-Chat + DF} &          &                & \bd{66.6 (+1.2)} & \bd{63.2 (+2.6)} & \bd{73.0 (+4.0)} & \bd{63.7 (+2.2)} \\
    \end{tabular}
    \vspace{-5pt}
\end{table*}

\section{Experiments}
\label{sec:experiments}

\subsection{Benchmarks}
To thoroughly assess our proposed DualFocus enhanced MLLM, we evaluated its performance across a spectrum of benchmarks, covering traditional academic Visual Question Answering (VQA) tasks (GQA \cite{hudson2018gqa}, TextVQA \cite{singh2019towards}) and recent benchmarks specifically designed for evaluating large multimodal models, namely MMBench \cite{MMBench} and SEED \cite{seed_2023}.
MMBench is constructed with manually designed questions to critically assess the model's vision-related reasoning and perceptual abilities.
SEED, leveraging GPT-4 for generation, introduces a dataset of nearly 19,000 questions centered on images and videos. Herein, our emphasis is placed on the image component., referred to as SEED$^{IMG}$.
GQA and TextVQA represent benchmarks in the domain of traditional Visual Question Answering tasks, with GQA assessing the model's ability to answer open-ended questions about images accurately, and TextVQA focusing on questions requiring the understanding of text within images.
Notably, GQA's evaluations revealed considerable variability due to discrepancies in the answer format. To address this, we employed GPT-3.5 to reformat answers into a multiple-choice question format, resulting in an adjusted benchmark referred to as GQA$^{*}$.

\subsection{Implementation Details}
All experiments were performed using LLaVA-1.5 \cite{llava1_5} and Qwen-VL-Chat \cite{bai2023qwen}, adhering to their default hyper-parameters and training configurations, unless stated otherwise. Our methodology uniquely altered the fine-tuning stage by incorporating the converted 143k VG data to fortify the MLLM with the DualFocus mechanism. 
For LLaVA-1.5, CLIP-ViT-L \cite{radford2021learning} served as the visual encoder at 336-resolution, and Vicuna 7B(13B) \cite{vicuna2023} functioned as the LLM. During training we only freeze the visual encoder but fine-tune the connection layers and LLM. 
For Qwen-VL-Chat, CLIP-ViT-G served as the visual encoder at 448-resolution and Qwen-7B \cite{bai2023qwen} functioned as its LLM. During training, given memory constraints, we freeze the visual encoder and LLM, only fine-tune the LoRA \cite{hu2022lora} and connection layers. 
The fine-tuning process for both models was completed within a single epoch.

\subsection{Main Results}
\bd{Comparison with Baseline Model}.
We first conducted comparisons against baseline MLLMs LLaVA-1.5 and Qwen-VL-Chat across four benchmarks: SEED, MMBench, GQA, and TextVQA. Our DualFocus mechanism notably enhances the performance of both methods, as outlined in Table \ref{tab:main_base}.
Specifically, our DualFocus improves LLaVA-1.5 with Vicuna-7B by 2.7, 2.3, 2.1, and 3.8, respectively. 
With the larger LLM, Vicuna-13B, DualFocus secures even more substantial gains: 2.8, 3.0, 4.2, and 4.2, on SEED, MMBench, GQA, and TextVQA, respectively. 
This trend is consistent when applying DualFocus to Qwen-VL-Chat, yielding boosts of 1.2, 2.6, 4.0 and 2.2 on the same benchmarks, respectively. These results highlight DualFocus's versatility and its capability to significantly elevate MLLM performance across diverse benchmarks.

\begin{table*}[h]
    \centering
    \caption{\bd{Comparison with SoTA methods on various benchmarks.} The best result and the second-best result should be indicated using bold and underline, respectively.}
    \label{tab:main_sota}
    \begin{tabular}{l|c|c|c|c|c|c|c}
    Method          & Res & Encoder-V & LLM         & SEED$^{IMG}$ & MMB & GQA$^{*}$ & TQA$^{T}$ \\
    \midrule
    InstructBLIP    & 224 & ViT-G     & Vicuna-7B   & 53.4 & 36.0 & -    & 50.1 \\
    LLaVA           & 224 & ViT-L     & Vicuna-7B   & 25.5 & 34.1 & -    & -    \\
    LLaVA-1.5       & 336 & ViT-L     & Vicuna-7B   & 66.2 & 64.3 & 67.2 & 58.2 \\
    Share4V         & 336 & ViT-L     & Vicuna-7B   & 69.7 & \underline{68.8} & 70.5 & 60.4 \\
    Qwen-VL-Chat    & 448 & ViT-G     & Qwen-7B     & 65.4 & 58.2 & 69.0 & 61.5 \\
    Monkey          & 896 & ViT-G     & Qwen-7B     & 64.3 & 59.6 & -    & \bd{67.6} \\
    OtterHD         & 1024 & -        & Fuyu-8B     & -    & 58.3 & -    & -    \\
    \midrule
    BLIP-2          & 224 & ViT-L     & Vicuna-13B  & -    & 46.4 & -    & 42.5 \\
    Shikra          & 224 & ViT-L     & Vicuna-13B  & -    & 58.8 & -    & - \\
    LLaVA-1.5       & 336 & ViT-L     & Vicuna-13B  & 68.2 & 67.7 & 69.3 & 61.3 \\
    Share4V         & 336 & ViT-L     & Vicuna-13B  & 70.8 & 68.5 & 71.1  & 62.2 \\
    \midrule
    \bd{LLaVA-1.5-DF (ours)} & 336 & ViT-L & Vicuna-13B  & \underline{71.0}     & \bd{70.7} & \underline{74.5}     & 65.5 \\
    \textbf{Share4V-DF (ours)}   & 336 & ViT-L & Vicuna-13B  & \bd{72.9} & \bd{70.7} & \bd{75.7} & \underline{66.2} \\
    \end{tabular}
    \vspace{-2mm}
\end{table*}

\bd{Comparison with SoTA Model.}
Subsequently, we conduct a comparison of DualFocus with other SoTA MLLMs that vary in their input resolutions (Res), visual encoders (Encoder-V), and language models (LLM) on Table \ref{tab:main_sota}.
We incorporate our DualFocus into LLaVA-1.5 Vicuna-13B and ShareGPT4V \cite{chen2023sharegpt4v}, a derivative of LLaVA, named as LLaVA-1.5-DF and Share4V-DF, exhibit superior performance across four distinct benchmarks.
Specifically, Share4V-DF surpasses its closest competitor by 2 on the SEED benchmark. Similarly, LLaVA-1.5-DF leads the second-best performer by 1.9 on the MMBench. 
The results are even more pronounced on the GQA and Text-VQA benchmarks, which demand a higher capacity for detailed perception.
Specifically, DualFocus improved Share4V by 4.7 and 4.0 on these two benchmarks, respectively.
While Monkey \cite{li2023monkey} achieves the highest score of 67.6 on TextVQA using a larger input of 896 x 896, it falls short on more comprehensive benchmarks like SEED and MMBench. In contrast, our Share4V-DF performs similarly on TextVQA with a much smaller input size of 336 x 336 and significantly better on the other two benchmarks, demonstrating DualFocus's ability to maintain a balance between a micro and macro perspective, making it a versatile and efficient mechanism for improving MLLM performance.

\subsection{\bd{Ablation Study}}
In this section, we first analyze the impact of each inference pathway, and then we show the effect of each component and why they work. Unless otherwise specified, all ablation results are based on LLaVA-1.5.

\begin{table}[t]
    \centering
    \caption{Performance comparison on different inference strategy for baseline LLaVA-1.5 and our model. ``Macro" and ``Micro" refers to employ macro and micro answer pathway, respectively. ``N/A" denotes the model fail to follow the instruction.}
    \label{tab:component}
    \begin{tabular}{c | c c|c c}
    Method                 & Macro      & Micro       & SEED$^I$ & VQA$^T$\\
    \midrule
    \multirow{2}{*}{Base}  & \checkmark &             & 66.2     & 58.2 \\
                           &            & \checkmark  & N/A      & N/A \\
    \midrule
    \multirow{3}{*}{Ours}  & \checkmark &             & 66.7 & 58.6 \\
                           &            & \checkmark  & 67.7 & 61.3  \\
                           & \checkmark & \checkmark  & \bd{68.9} & \bd{62.0} \\
    \end{tabular}
    \vspace{-2mm}
\end{table}

\bd{Inference Pathway Analysis}.
Table \ref{tab:component} illustrates the contributions of the micro and macro inference pathways to our method's performance. The initial results from the baseline model, LLaVA-1.5, indicate failure to implement the micro pathway due to the absence of training with similar directives. The integration of our custom 143k VG dataset, enabled the model to follow the DF inference guidelines. However, this adaptation led to minor improvements, \ie, increasing by +0.5 on the SEED metric and +0.4 on the TextVQA metric, suggesting that the dataset alone is insufficient to enhance performance.

Furthermore, the micro pathway's implementation resulted in a significant +1.0 gain on the SEED metric and a notable +2.7 gain on the TextVQA metric, supporting our hypothesis that the micro pathway excels in nuanced tasks. Conversely, global comprehension tasks benefit from the PPL model, as evidenced by a +1.2 gain on the SEED metric and a moderate +0.7 gain on the TextVQA metric. This data underscores the importance of employing the appropriate inference pathway based on the task's requirements.

\begin{table}[t]
    \centering
    \caption{Results on POPE. ``LLaVA" refers to LLaVA-1.5. DualFocus is beneficial to mitigate Hallucination of MLLM. Here A, P, R denotes adversarial, popular and random split of POPE, respectively. ``F1" and ``Acc" denotes F1 score and accuracy, respectively.}
    \label{tab:pope}
    \resizebox{\linewidth}{!}{
    \begin{tabular}{l|c|c|c|c|c|c}
         & F1(A) & Acc(A) & F1(P) & Acc(P) & F1(R) & Acc(R) \\
         \midrule
    LLaVA        & 84.2 & 85.2 & 86.2 & 87.3 & 87.4 & 88.2 \\
    LLaVA + DF   & \bd{86.0} & \bd{86.2} & \bd{88.6} & \bd{89.1} & \bd{89.7} & \bd{90.0} \\
    \end{tabular}}
    \vspace{-3mm}
\end{table}

\begin{figure}[t]
    \centering
    \includegraphics[width=\linewidth]{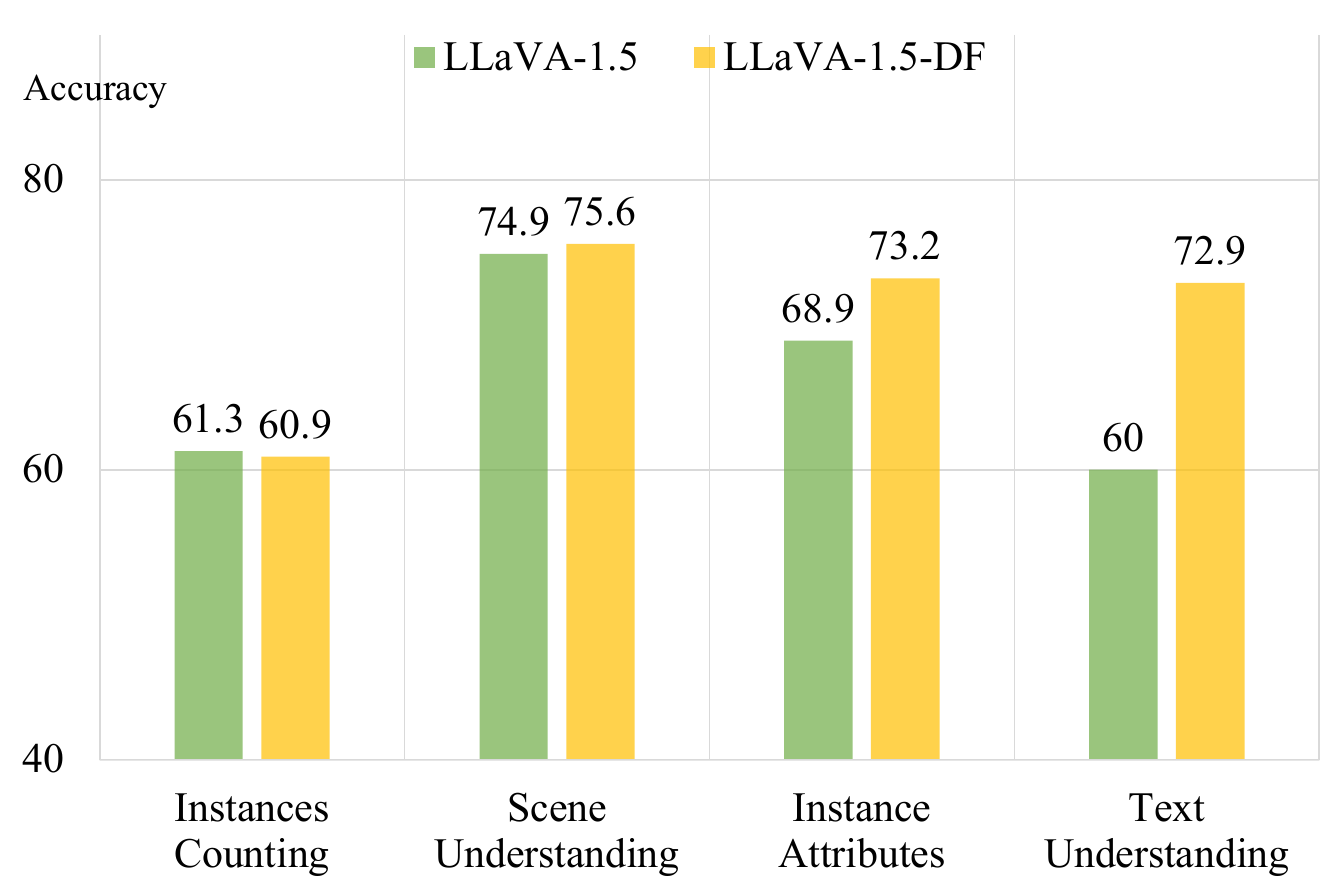}
    \caption{Accuracy of baseline LLaVA-1.5 and our LLaVA-1.5-DF on SEED Benchmark tasks across various granularities. Our Dual-Focus significantly improves accuracy on fine-grained tasks.}
    \label{fig:seed-fine}
\vspace{-4mm}
\end{figure}

\begin{figure*}[ht]
    \centering
    \includegraphics[width=\linewidth]{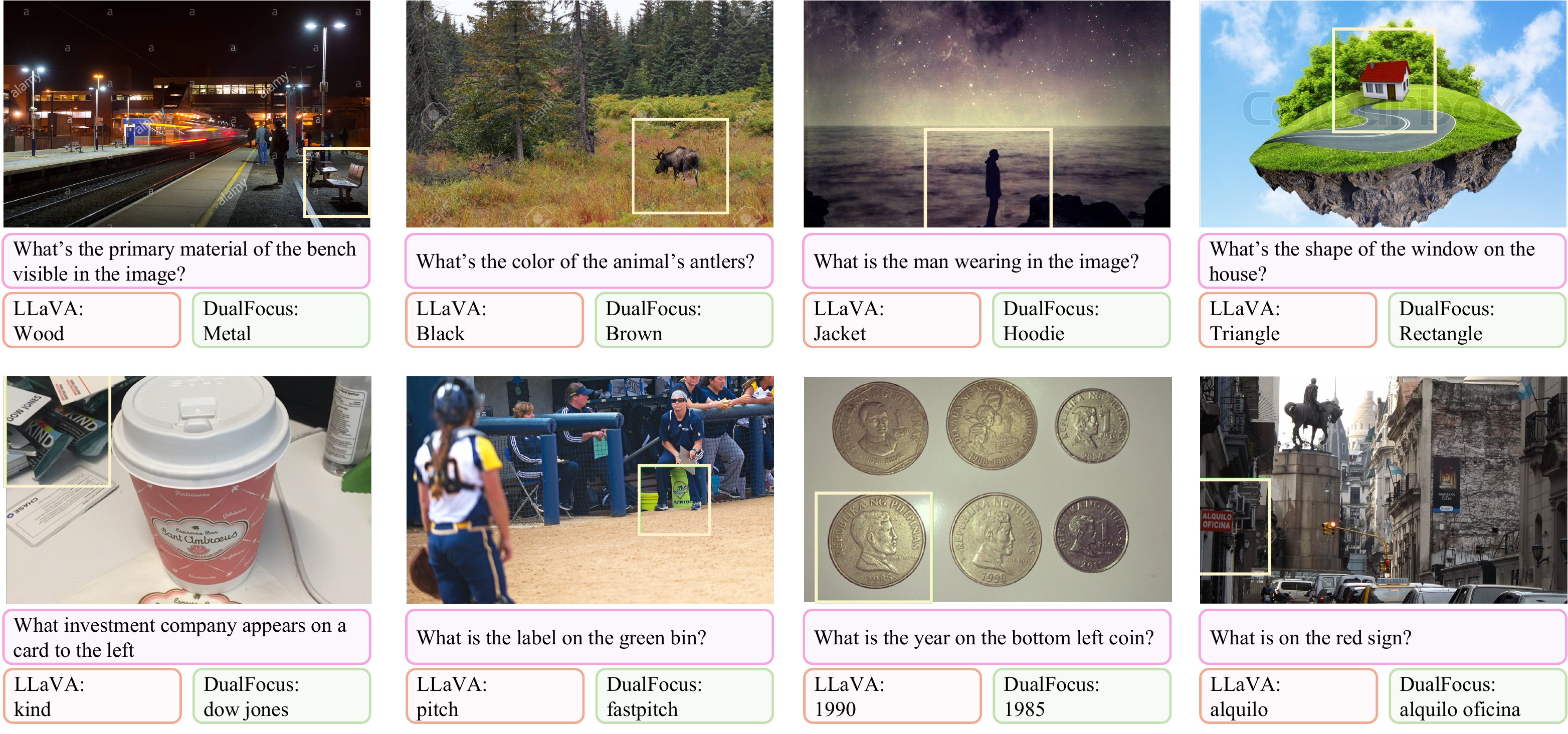}
    \caption{Comparative visualizations between LLaVA-1.5 and our DualFocus method on SEED (first row) and TextVQA datasets (second row). LLaVA-1.5 is struggle to capture micro details. In contrast, our DualFocus mechanism leverages the zoomed-in sub-region (highlighted with a yellow bounding box) to achieve improved discernment of fine-grained details.}
    \label{fig:Demo}
    \vspace{-4mm}
\end{figure*}

\bd{Hallucination Mitigation}.
Hallucination within MLLM presents a critical challenge where the model creates imaginary content that is not actually present in the image. The benchmark POPE \cite{li2023evaluating}, is designed to evaluate such hallucinations in MLLM through three distinct data splits: adversarial (A), popular (P), and random (R). As indicated in Table \ref{tab:pope}, integrating our DualFocus into MLLM yields substantial improvements in accuracy and the F1 score across these data splits.
Specifically, it improvements baseline by 2.4 and 2.3 on the F1 score of splits ``P" and ``R", respectively. Even on the most difficult split ``A", it yields 1.8 gains on the F1 score.
The effectiveness of DF is attributed to that our DualFocus directs the model’s attention towards specific, relevant parts of an image in connection to the posed question, reducing the generation of non-pertinent features and subsequently diminishing the likelihood of hallucinations.

\bd{Fine-Grained Perception Enhancement}.
In this section, we delve into the effectiveness of the DualFocus mechanism on tasks emphasizing different cognitive demands. 
For a clear demonstration, we only involve the micro answer pathway here.
We use SEED as a benchmark because it provides a comprehensive assessment of a model's capabilities across different dimensions and levels of detail. 
Specifically, we examine four key dimensions: Instance Counting, Scene Understanding, Instance Attributes, and Text Understanding. The first two dimensions primarily concern the broader context of a situation, emphasizing a macro perspective. In contrast, the latter two focus on more intricate, micro-level details.
Experiment results are presented in Figure \ref{fig:seed-fine}, illustrating that while our DualFocus mechanism delivers modest improvements in the domains of Instances Counting and Scene Understanding (-0.4, +0.7), it significantly enhances performance on Instance Attributes and Text Understanding (+4.3, +12.9). 
These results underscore the effectiveness of the DualFocus approach, particularly in tasks requiring acute attention to detail, thereby confirming its utility in dissecting and interpreting finer elements within data.

\begin{table}[t]
    \centering
    \caption{Performance comparison of different methods for implementing the PPL strategy on MMBench. 
    ``LLaVA" and ``Qwen" refer to LLaVA-1.5 and Qwen-VL-Chat. 
    ``LLaVA + PPL" denotes using PPL to choose answers generated by LLaVA with two distinct prompts, a process mirrored in ``Qwen + PPL".
    ``LLaVA + Qwen + PPL" refer to use PPL to select the best answer from LLaVA and Qwen.
    }
    \label{tab:ppl}
    \tablestyle{6pt}{1.2}
    \begin{tabular}{c|c|c|c|c}
    Method                & LLaVA      & Qwen       & PPL           & Acc \\
    \midrule
    \multirow{5}{*}{Base} & \checkmark &            &               & 64.3 \\
                          & \checkmark &            & \checkmark    & 64.1 \\
                          &            & \checkmark &               & 60.6 \\
                          &            & \checkmark & \checkmark    & 60.7 \\
                          & \checkmark & \checkmark & \checkmark    & 62.5 \\
    \midrule
    \midrule
    \multirow{2}{*}{Our}  &   \multicolumn{3}{c|}{LLaVA-DF}         & 66.6 \\          
                &    \multicolumn{3}{c|}{Qwen-DF}                   & 63.2 \\    
    \midrule
    \end{tabular}
    \vspace{-5mm}
\end{table}

\bd{PPL-Guided Answer Specialization}.
During inference, we utilized PPL to select answers by combining the micro and macro inference pathways, essentially creating a unique assembly method. We further examine various assembly approaches, detailed in Table \ref{tab:ppl}. 
The first strategy involves using PPL to merge answers from same models but varying input formats, labeled as ``LLaVA + PPL" and ``Qwen + PPL".
Given that the base model is limited to macro inference pathways, we applied two distinct prompts. Results indicate a minor impact on performance, with changes of -0.2 and +0.1, respectively.
Another assembly strategy involves using PPL to merge answers from different models, tagged as ``LLaVA + Qwen + PPL". This approach significantly improved Qwen by +1.9, yet it reduced LLaVA’s performance by 1.8. We suspect this variance results from differing model architectures and training methodologies. 
In contrast, DualFocus integrates micro and macro pathways within each model, applied to LLaVA and Qwen resulted in substantial gains of +2.3 and +2.6, respectively, are both higher than ``LLaVA + Qwen + PPL". This suggests combining micro and macro inferences within a single model outperforms assembling answers across different models.

\vspace{-2mm}

\section{Conclusion}
\label{sec:conclusion}

In this work, we introduced DualFocus, a novel approach to enhance the performance of Multi-modal Large Language Models (MLLMs) by integrating both macro and micro perspectives for improved visual question answering. Through comparative studies, DualFocus demonstrated superior capability in handling detailed features and mitigating hallucination, thereby outperforming existing methods. This method not only advances MLLM efficacy but also paves the way for more human-like visual reasoning in AI.
\newpage

\bibliography{example_paper}

\begin{thebibliography}{47}
\providecommand{\natexlab}[1]{#1}
\providecommand{\url}[1]{\texttt{#1}}
\expandafter\ifx\csname urlstyle\endcsname\relax
  \providecommand{\doi}[1]{doi: #1}\else
  \providecommand{\doi}{doi: \begingroup \urlstyle{rm}\Url}\fi

\bibitem[Bai et~al.(2023)Bai, Bai, Yang, Wang, Tan, Wang, Lin, Zhou, and Zhou]{bai2023qwen}
Bai, J., Bai, S., Yang, S., Wang, S., Tan, S., Wang, P., Lin, J., Zhou, C., and Zhou, J.
\newblock Qwen-vl: A frontier large vision-language model with versatile abilities.
\newblock \emph{arXiv.org}, 2023.

\bibitem[Baichuan(2023)]{baichuan2023baichuan2}
Baichuan.
\newblock Baichuan 2: Open large-scale language models.
\newblock \emph{arXiv.org}, 2023.
\newblock URL \url{https://arxiv.org/abs/2309.10305}.

\bibitem[Brown et~al.(2020)Brown, Mann, Ryder, Subbiah, Kaplan, Dhariwal, Neelakantan, Shyam, Sastry, Askell, et~al.]{brown2020language}
Brown, T., Mann, B., Ryder, N., Subbiah, M., Kaplan, J.~D., Dhariwal, P., Neelakantan, A., Shyam, P., Sastry, G., Askell, A., et~al.
\newblock Language models are few-shot learners.
\newblock \emph{Advances in Neural Information Processing Systems (NeurIPS)}, 33:\penalty0 1877--1901, 2020.

\bibitem[Chen et~al.(2023{\natexlab{a}})Chen, Zhang, Zeng, Zhang, Zhu, and Zhao]{chen2023shikra}
Chen, K., Zhang, Z., Zeng, W., Zhang, R., Zhu, F., and Zhao, R.
\newblock Shikra: Unleashing multimodal llm's referential dialogue magic.
\newblock \emph{arXiv.org}, 2023{\natexlab{a}}.

\bibitem[Chen et~al.(2023{\natexlab{b}})Chen, Li, Dong, Zhang, He, Wang, Zhao, and Lin]{chen2023sharegpt4v}
Chen, L., Li, J., Dong, X., Zhang, P., He, C., Wang, J., Zhao, F., and Lin, D.
\newblock Sharegpt4v: Improving large multi-modal models with better captions.
\newblock \emph{arXiv.org}, 2023{\natexlab{b}}.

\bibitem[Chiang et~al.(2023)Chiang, Li, Lin, Sheng, Wu, Zhang, Zheng, Zhuang, Zhuang, Gonzalez, Stoica, and Xing]{vicuna2023}
Chiang, W.-L., Li, Z., Lin, Z., Sheng, Y., Wu, Z., Zhang, H., Zheng, L., Zhuang, S., Zhuang, Y., Gonzalez, J.~E., Stoica, I., and Xing, E.~P.
\newblock Vicuna: An open-source chatbot impressing gpt-4 with 90\%* chatgpt quality, March 2023.
\newblock URL \url{https://lmsys.org/blog/2023-03-30-vicuna/}.

\bibitem[Chowdhery et~al.(2022)Chowdhery, Narang, Devlin, Bosma, Mishra, Roberts, Barham, Chung, Sutton, Gehrmann, et~al.]{chowdhery2022palm}
Chowdhery, A., Narang, S., Devlin, J., Bosma, M., Mishra, G., Roberts, A., Barham, P., Chung, H.~W., Sutton, C., Gehrmann, S., et~al.
\newblock Palm: Scaling language modeling with pathways.
\newblock \emph{arXiv.org}, 2022.

\bibitem[Dai et~al.(2023)Dai, Li, Li, Tiong, Zhao, Wang, Li, Fung, and Hoi]{dai2023instructblip}
Dai, W., Li, J., Li, D., Tiong, A. M.~H., Zhao, J., Wang, W., Li, B., Fung, P., and Hoi, S.
\newblock Instructblip: Towards general-purpose vision-language models with instruction tuning, 2023.

\bibitem[Devlin et~al.(2018)Devlin, Chang, Lee, and Toutanova]{devlin2018bert}
Devlin, J., Chang, M.-W., Lee, K., and Toutanova, K.
\newblock Bert: Pre-training of deep bidirectional transformers for language understanding.
\newblock \emph{arXiv.org}, 2018.

\bibitem[Dong et~al.(2024)Dong, Zhang, Zang, Cao, Wang, Ouyang, Wei, Zhang, Duan, Cao, Zhang, Li, Yan, Gao, Zhang, Li, Li, Chen, He, Zhang, Qiao, Lin, and Wang]{internlmxcomposer2}
Dong, X., Zhang, P., Zang, Y., Cao, Y., Wang, B., Ouyang, L., Wei, X., Zhang, S., Duan, H., Cao, M., Zhang, W., Li, Y., Yan, H., Gao, Y., Zhang, X., Li, W., Li, J., Chen, K., He, C., Zhang, X., Qiao, Y., Lin, D., and Wang, J.
\newblock Internlm-xcomposer2: Mastering free-form text-image composition and comprehension in vision-language large model.
\newblock \emph{arXiv.org}, 2024.

\bibitem[Ge et~al.()Ge, Ge, Zeng, Wang, and Shan]{seed_2023}
Ge, Y., Ge, Y., Zeng, Z., Wang, X., and Shan, Y.
\newblock Planting a seed of vision in large language model.

\bibitem[He et~al.(2023)He, Jin, Xu, Qiu, Wang, Li, Yan, Wang, and Lin]{He2023WanJuanAC}
He, C., Jin, Z., Xu, C., Qiu, J., Wang, B., Li, W., Yan, H., Wang, J., and Lin, D.
\newblock Wanjuan: A comprehensive multimodal dataset for advancing english and chinese large models.
\newblock \emph{arXiv.org}, abs/2308.10755, 2023.
\newblock URL \url{https://api.semanticscholar.org/CorpusID:261049100}.

\bibitem[Hong et~al.(2023)Hong, Wang, Lv, Xu, Yu, Ji, Wang, Wang, Dong, Ding, et~al.]{hong2023cogagent}
Hong, W., Wang, W., Lv, Q., Xu, J., Yu, W., Ji, J., Wang, Y., Wang, Z., Dong, Y., Ding, M., et~al.
\newblock Cogagent: A visual language model for gui agents.
\newblock \emph{arXiv.org}, 2023.

\bibitem[Hu et~al.(2022)Hu, Shen, Wallis, Allen-Zhu, Li, Wang, Wang, and Chen]{hu2022lora}
Hu, E.~J., Shen, Y., Wallis, P., Allen-Zhu, Z., Li, Y., Wang, S., Wang, L., and Chen, W.
\newblock Lo{RA}: Low-rank adaptation of large language models.
\newblock In \emph{International Conference on Learning Representations}, 2022.
\newblock URL \url{https://openreview.net/forum?id=nZeVKeeFYf9}.

\bibitem[Hu et~al.(2023)Hu, Xu, Li, Li, Chen, and Tu]{Hu2023BLIVAAS}
Hu, W., Xu, Y., Li, Y., Li, W., Chen, Z., and Tu, Z.
\newblock Bliva: A simple multimodal llm for better handling of text-rich visual questions.
\newblock \emph{ArXiv}, abs/2308.09936, 2023.
\newblock URL \url{https://api.semanticscholar.org/CorpusID:261049015}.

\bibitem[Huang et~al.(2023)Huang, Dong, Zhang, Wang, He, Wang, Lin, Zhang, and Yu]{huang2023opera}
Huang, Q., Dong, X., Zhang, P., Wang, B., He, C., Wang, J., Lin, D., Zhang, W., and Yu, N.
\newblock Opera: Alleviating hallucination in multi-modal large language models via over-trust penalty and retrospection-allocation.
\newblock \emph{arXiv.org}, 2023.

\bibitem[Hudson \& Manning(2019)Hudson and Manning]{hudson2018gqa}
Hudson, D.~A. and Manning, C.~D.
\newblock Gqa: A new dataset for real-world visual reasoning and compositional question answering.
\newblock \emph{Conference on Computer Vision and Pattern Recognition (CVPR)}, 2019.

\bibitem[Jelinek(1998)]{jelinek1998statistical}
Jelinek, F.
\newblock \emph{Statistical methods for speech recognition}.
\newblock MIT press, 1998.

\bibitem[Krishna et~al.(2017)Krishna, Zhu, Groth, Johnson, Hata, Kravitz, Chen, Kalantidis, Li, Shamma, et~al.]{krishna2017visual}
Krishna, R., Zhu, Y., Groth, O., Johnson, J., Hata, K., Kravitz, J., Chen, S., Kalantidis, Y., Li, L.-J., Shamma, D.~A., et~al.
\newblock Visual genome: Connecting language and vision using crowdsourced dense image annotations.
\newblock \emph{IJCV}, 2017.

\bibitem[Li et~al.(2023{\natexlab{a}})Li, Zhang, Yang, Zhang, Pu, and Liu]{li2023otterhd}
Li, B., Zhang, P., Yang, J., Zhang, Y., Pu, F., and Liu, Z.
\newblock Otterhd: A high-resolution multi-modality model.
\newblock \emph{Arxiv}, 2023{\natexlab{a}}.

\bibitem[Li et~al.(2023{\natexlab{b}})Li, Zhang, Chen, Wang, Yang, and Liu]{li2023otter}
Li, B., Zhang, Y., Chen, L., Wang, J., Yang, J., and Liu, Z.
\newblock Otter: A multi-modal model with in-context instruction tuning.
\newblock \emph{arXiv.org}, 2023{\natexlab{b}}.

\bibitem[Li et~al.(2022)Li, Li, Xiong, and Hoi]{li2022blip}
Li, J., Li, D., Xiong, C., and Hoi, S.
\newblock Blip: Bootstrapping language-image pre-training for unified vision-language understanding and generation.
\newblock In \emph{Proceedings of the International Conference on Machine learning (ICML)}, pp.\  12888--12900. PMLR, 2022.

\bibitem[Li et~al.(2023{\natexlab{c}})Li, Du, Zhou, Wang, Zhao, and Wen]{li2023evaluating}
Li, Y., Du, Y., Zhou, K., Wang, J., Zhao, W.~X., and Wen, J.-R.
\newblock Evaluating object hallucination in large vision-language models.
\newblock \emph{arXiv.org}, 2023{\natexlab{c}}.

\bibitem[Li et~al.(2023{\natexlab{d}})Li, Yang, Liu, Ma, Zhang, Yang, Sun, Liu, and Bai]{li2023monkey}
Li, Z., Yang, B., Liu, Q., Ma, Z., Zhang, S., Yang, J., Sun, Y., Liu, Y., and Bai, X.
\newblock Monkey: Image resolution and text label are important things for large multi-modal models.
\newblock \emph{Arxiv}, 2023{\natexlab{d}}.

\bibitem[Liu et~al.(2023{\natexlab{a}})Liu, Li, Li, and Lee]{llava1_5}
Liu, H., Li, C., Li, Y., and Lee, Y.~J.
\newblock Improved baselines with visual instruction tuning.
\newblock \emph{arXiv preprint arXiv:2310.03744}, 2023{\natexlab{a}}.

\bibitem[Liu et~al.(2023{\natexlab{b}})Liu, Li, Wu, and Lee]{liu2023visual}
Liu, H., Li, C., Wu, Q., and Lee, Y.~J.
\newblock Visual instruction tuning.
\newblock \emph{arXiv.org}, 2023{\natexlab{b}}.

\bibitem[Liu et~al.(2023{\natexlab{c}})Liu, Duan, Zhang, Li, Zhnag, Zhao, Yuan, Wang, He, Liu, Chen, and Lin]{MMBench}
Liu, Y., Duan, H., Zhang, Y., Li, B., Zhnag, S., Zhao, W., Yuan, Y., Wang, J., He, C., Liu, Z., Chen, K., and Lin, D.
\newblock Mmbench: Is your multi-modal model an all-around player?
\newblock \emph{arXiv:2307.06281}, 2023{\natexlab{c}}.

\bibitem[OpenAI(2022)]{openai2020chatgpt}
OpenAI.
\newblock Chatgpt.
\newblock \url{https://openai.com/blog/chatgpt}, 2022.

\bibitem[OpenAI(2023)]{openai2023gpt4}
OpenAI.
\newblock Gpt-4 technical report, 2023.

\bibitem[Ouyang et~al.(2022)Ouyang, Wu, Jiang, Almeida, Wainwright, Mishkin, Zhang, Agarwal, Slama, Ray, et~al.]{ouyang2022training}
Ouyang, L., Wu, J., Jiang, X., Almeida, D., Wainwright, C., Mishkin, P., Zhang, C., Agarwal, S., Slama, K., Ray, A., et~al.
\newblock Training language models to follow instructions with human feedback.
\newblock \emph{Advances in Neural Information Processing Systems (NeurIPS)}, 35:\penalty0 27730--27744, 2022.

\bibitem[Peng et~al.(2023)Peng, Wang, Dong, Hao, Huang, Ma, and Wei]{peng2023kosmos}
Peng, Z., Wang, W., Dong, L., Hao, Y., Huang, S., Ma, S., and Wei, F.
\newblock Kosmos-2: Grounding multimodal large language models to the world.
\newblock \emph{arXiv.org}, 2023.

\bibitem[Qi et~al.(2023{\natexlab{a}})Qi, Fang, Sun, Wu, Wu, Wang, Lin, and Zhao]{qi2023gpt4point}
Qi, Z., Fang, Y., Sun, Z., Wu, X., Wu, T., Wang, J., Lin, D., and Zhao, H.
\newblock Gpt4point: A unified framework for point-language understanding and generation, 2023{\natexlab{a}}.

\bibitem[Qi et~al.(2023{\natexlab{b}})Qi, Fang, Zhang, Sun, Wu, Liu, Lin, Wang, and Zhao]{qi2023gemini}
Qi, Z., Fang, Y., Zhang, M., Sun, Z., Wu, T., Liu, Z., Lin, D., Wang, J., and Zhao, H.
\newblock Gemini vs gpt-4v: A preliminary comparison and combination of vision-language models through qualitative cases, 2023{\natexlab{b}}.

\bibitem[Qwen(2023)]{qwen7b}
Qwen.
\newblock Introducing qwen-7b: Open foundation and human-aligned models (of the state-of-the-arts), 2023.

\bibitem[Radford et~al.(2021)Radford, Kim, Hallacy, Ramesh, Goh, Agarwal, Sastry, Askell, Mishkin, Clark, et~al.]{radford2021learning}
Radford, A., Kim, J.~W., Hallacy, C., Ramesh, A., Goh, G., Agarwal, S., Sastry, G., Askell, A., Mishkin, P., Clark, J., et~al.
\newblock Learning transferable visual models from natural language supervision.
\newblock In \emph{Proceedings of the International Conference on Machine learning (ICML)}, pp.\  8748--8763. PMLR, 2021.

\bibitem[Raffel et~al.(2020)Raffel, Shazeer, Roberts, Lee, Narang, Matena, Zhou, Li, and Liu]{raffel2020exploring}
Raffel, C., Shazeer, N., Roberts, A., Lee, K., Narang, S., Matena, M., Zhou, Y., Li, W., and Liu, P.~J.
\newblock Exploring the limits of transfer learning with a unified text-to-text transformer.
\newblock \emph{Journal of Machine Learning Research (JMLR)}, 21\penalty0 (1):\penalty0 5485--5551, 2020.

\bibitem[Singh et~al.(2019)Singh, Natarajan, Shah, Jiang, Chen, Batra, Parikh, and Rohrbach]{singh2019towards}
Singh, A., Natarajan, V., Shah, M., Jiang, Y., Chen, X., Batra, D., Parikh, D., and Rohrbach, M.
\newblock Towards vqa models that can read.
\newblock In \emph{Proceedings of the IEEE/CVF conference on computer vision and pattern recognition}, pp.\  8317--8326, 2019.

\bibitem[Sun et~al.(2023)Sun, Fang, Wu, Zhang, Zang, Kong, Xiong, Lin, and Wang]{sun2023alpha}
Sun, Z., Fang, Y., Wu, T., Zhang, P., Zang, Y., Kong, S., Xiong, Y., Lin, D., and Wang, J.
\newblock {Alpha-CLIP}: A clip model focusing on wherever you want.
\newblock \emph{arXiv.org}, 2023.

\bibitem[Team(2023)]{2023internlm}
Team, I.
\newblock Internlm: A multilingual language model with progressively enhanced capabilities.
\newblock \url{https://github.com/InternLM/InternLM}, 2023.

\bibitem[Touvron et~al.(2023{\natexlab{a}})Touvron, Lavril, Izacard, Martinet, Lachaux, Lacroix, Rozi{\`e}re, Goyal, Hambro, Azhar, et~al.]{touvron2023llama}
Touvron, H., Lavril, T., Izacard, G., Martinet, X., Lachaux, M.-A., Lacroix, T., Rozi{\`e}re, B., Goyal, N., Hambro, E., Azhar, F., et~al.
\newblock Llama: Open and efficient foundation language models.
\newblock \emph{arXiv.org}, 2023{\natexlab{a}}.

\bibitem[Touvron et~al.(2023{\natexlab{b}})Touvron, Martin, Stone, Albert, Almahairi, Babaei, Bashlykov, Batra, Bhargava, Bhosale, et~al.]{touvron2023llama2}
Touvron, H., Martin, L., Stone, K., Albert, P., Almahairi, A., Babaei, Y., Bashlykov, N., Batra, S., Bhargava, P., Bhosale, S., et~al.
\newblock Llama 2: Open foundation and fine-tuned chat models, 2023{\natexlab{b}}.

\bibitem[Wang et~al.(2023)Wang, Zhang, Chu, Cao, Zhou, Wu, Wang, He, and Lin]{wang2023v3det}
Wang, J., Zhang, P., Chu, T., Cao, Y., Zhou, Y., Wu, T., Wang, B., He, C., and Lin, D.
\newblock V3det: Vast vocabulary visual detection dataset.
\newblock In \emph{Proceedings of the IEEE/CVF International Conference on Computer Vision (ICCV)}, October 2023.

\bibitem[Wei et~al.(2022)Wei, Wang, Schuurmans, Bosma, Xia, Chi, Le, Zhou, et~al.]{wei2022chain}
Wei, J., Wang, X., Schuurmans, D., Bosma, M., Xia, F., Chi, E., Le, Q.~V., Zhou, D., et~al.
\newblock Chain-of-thought prompting elicits reasoning in large language models.
\newblock \emph{Advances in Neural Information Processing Systems (NIPS)}, 35:\penalty0 24824--24837, 2022.

\bibitem[Ye et~al.(2023)Ye, Xu, Xu, Ye, Yan, Zhou, Wang, Hu, Shi, Shi, et~al.]{ye2023mplug}
Ye, Q., Xu, H., Xu, G., Ye, J., Yan, M., Zhou, Y., Wang, J., Hu, A., Shi, P., Shi, Y., et~al.
\newblock mplug-owl: Modularization empowers large language models with multimodality.
\newblock \emph{arXiv.org}, 2023.

\bibitem[Zhang et~al.(2023)Zhang, Dong, Wang, Cao, Xu, Ouyang, Zhao, Ding, Zhang, Duan, Zhang, Yan, Zhang, Li, Li, Chen, He, Zhang, Qiao, Lin, and Wang]{internlmxcomposer}
Zhang, P., Dong, X., Wang, B., Cao, Y., Xu, C., Ouyang, L., Zhao, Z., Ding, S., Zhang, S., Duan, H., Zhang, W., Yan, H., Zhang, X., Li, W., Li, J., Chen, K., He, C., Zhang, X., Qiao, Y., Lin, D., and Wang, J.
\newblock Internlm-xcomposer: A vision-language large model for advanced text-image comprehension and composition.
\newblock \emph{arXiv.org}, 2023.

\bibitem[Zhao et~al.(2023)Zhao, Ouyang, Wang, Huang, Zhang, Dong, Wang, and He]{zhao2023mllm}
Zhao, Z., Ouyang, L., Wang, B., Huang, S., Zhang, P., Dong, X., Wang, J., and He, C.
\newblock Mllm-dataengine: An iterative refinement approach for mllm.
\newblock \emph{arXiv.org}, 2023.

\bibitem[Zhu et~al.(2023)Zhu, Chen, Shen, Li, and Elhoseiny]{zhu2023minigpt}
Zhu, D., Chen, J., Shen, X., Li, X., and Elhoseiny, M.
\newblock Minigpt-4: Enhancing vision-language understanding with advanced large language models.
\newblock \emph{arXiv.org}, 2023.

\end{thebibliography}
\bibliographystyle{icml2024}


%

\end{document}